# Efficient Algorithms for Parsing the DOP Model ?
# A Reply to Joshua Goodman

**Rens Bod**
University of Amsterdam
Department of Computational Linguistics
Spuistraat 134, NL-1012 VB Amsterdam
Rens.Bod@let.uva.nl

This note is a reply to Joshua Goodman's paper "Efficient Algorithms for Parsing the DOP Model" (Goodman, 1996). In his paper, Goodman makes a number of claims about (my work on) the Data-Oriented Parsing model (Bod, 1992-1996). This note shows that some of these claims must be mistaken.

## 1.    Goodman did not find an efficient algorithm for "DOP"

Goodman claims to have found an efficient polynomial algorithm for parsing the DOP model. However, Goodman's model generates best parse trees that can never be produced by the DOP model.

In the introduction of his paper, Goodman observes that the DOP model ".. can be summarized as a special kind of Stochastic Tree Substitution Grammar (STSG): given a bracketed, labeled training corpus, let *every* subtree of that corpus be an elementary tree, with a probability proportional to the number of occurrences of that subtree in the training corpus." Goodman then neglects to add that according to the DOP model, the "preferred" or "best" parse tree of a sentence is the most probable parse tree of that sentence. This definition is found in all my publications about DOP (e.g. Bod, 1992-96; van den Berg, Bod and Scha, 1994; Bod and Scha, 1994; Bod, Krauwer and Sima'an, 1994; Bod, Bonnema and Scha, 1996; Sima'an, Bod, Krauwer and Scha, 1994). Since in DOP, the probability of a tree is the sum of the probabilities of all distinct derivations that produce that tree, the computation of the most probable tree is very expensive. In Sima'an (1996b), a proof is given that the problem of computing the most probable tree of a sentence in DOP is NP-hard. This proof does not mean that there is no algorithm that can *estimate* the most probable tree of a sentence with an error that can be made arbitrarily small (cf. Bod, 1993b), but it does mean that there is no *deterministic* polynomial time algorithm for finding the most probable tree.

Is then Goodman's claim that he found an efficient polynomial time algorithm for DOP wrong? Yes. Unless he means something different with "DOP". And this seems to be the case. In his section "Reduction of DOP to PCFG", Goodman states that he found a reduction of DOP to a probabilistic context-free grammar (PCFG) which ".. generates the same trees with the same probabilities, although one must sum over several PCFG trees for each STSG tree." But this reduction does still not allow for an efficient calculation of the most probable tree of a sentence in DOP or STSG.[1] However, Goodman shows that with his PCFG-reduction he can efficiently calculate the so-called "Maximum Constituents Parse", i.e. the parse tree which is most likely to have the largest number of correct constituents. He then shows in the section "Parsing Algorithm" that this Maximum Constituents Parse can produce trees that cannot be produced by the grammar (i.e. by the STSG underlying the DOP model). Thus, Goodman's model does *not* produce the same best trees as the DOP model. In other words, Goodman's model is not a DOP model.[2]

Note that Goodman's model is neither a special case of the DOP model, since it returns best trees that the DOP model could not even produce. Instantiations of DOP can be obtained by for instance constraining the size of the elementary trees (e.g. Charniak, 1996), or by estimating the most probable tree by the most probable derivation (e.g. Sima'an, 1995).

## 2.     Goodman's analysis of Bod's data is mistaken

Goodman analyzes that there is an upper bound on the probability of getting the test set Bod describes. His analysis is mistaken.

In his section "Analysis of Bod's Data", Goodman observes that ".. in the DOP model, a sentence cannot be given an exactly correct parse unless all productions in the correct parse occur in the training set." Although he refers here only to the model which I have called "DOP1" (Bod, 1995a), he states that this observation means that one can find ".. an upper bound on the probability that any particular level of performance could be achieved." He then gives a calculation showing that in the most generous case for a random split into test and training ".. there is only about a 1.5% chance of getting the test set Bod describes." Goodman's calculation is based on his following statement: "According to his thesis (Bod, 1995a, page 64), only one of his 75 test sentences had a correct parse which could not be generated from the training data." This statement is incorrect. In my thesis (page 64), I write: "It may be relevant to mention that the parse *coverage* was 99%." The notion of parse *coverage* is well-known in the literature (e.g. Grishman, 1989; Karlsson et al., 1995), and refers to the percentage of test sentences that could be parsed by a system/grammar. The *coverage* of a system should not be confused with its *structural consistency* (Black et al., 1993; Black, Garside and Leech, 1993), which is the percentage of test sentences for which a parse was found that is *equal* to the test set parse; (also called *correctness* of a system, cf. Karlsson et al., 1995). Now it appears that Goodman interpreted my reported 99% *coverage* as 99% *structural consistency*, which is a mistake.[3] Therefore Goodman's claim that ".. there is only about a 1.5% chance of getting the test set Bod describes" is also mistaken. Moreover, his claim that my results are "... due to an extremely fortuitous choice of test data" does therefore not hold.

---

[1] The fact that Goodman calls his reduction a "PCFG" is somewhat misleading. As is well-known, for a PCFG the most probable tree can be efficiently calculated by the Viterbi algorithm (cf. Jelinek et al., 1990).

[2] At the conference presentation of his paper, however, Goodman reported that he also reimplemented my Monte Carlo parser for estimating the most probable tree in DOP (cf. Bod, 1993-95). Unfortunately, he did not include this in his paper.

It should be stressed that it is not at all unlikely to get a random test set with a *coverage* of 99%; Khalil Sima'an consistently achieved 98-99% coverage on a large number of different random test sets (Sima'an, 1995, 1996a).

However, it may still be that the chance of getting a test set with my reported 96% parse *accuracy* (which is the percentage of the most probable parses that match with the test set parses) is small. In order to check this, I calculated the chance of getting a test set for which at least 96% of the sentences could be parsed correctly. According to my data (which seem to differ from the data Goodman says to have used, in that I *do* take into account epsilon productions, see Bod, 1995a, pp. 66-7, 77-8, 98), this chance is about 45%. This shows that my results are not due to an "extremely fortuitous choice of test data" as Goodman writes. Even if Goodman's "undergeneration probabilities" are used (derived from his data in table 4, Goodman 1996), the probability of getting a test set for which at least 96% of the sentences could be parsed correctly is about 14.5% in the most generous case.

DOP's high accuracy results were certainly due to using cleaner data than others, but also to using a richer statistical model than others (cf. Rajman, 1995). On unedited data from Penn's ATIS, DOP achieved 64% parse accuracy and 94.1% crossing brackets accuracy (Bod, 1996a).

## 3.  Bod's results are not due to a fortuitous choice of test data

This has been shown in the previous section.

## 4.  Final Remark: DOP outperforms other models

The DOP model, and all reported instantiations of it, outperform all other models it has been compared to (cf. Bod, 1996b; Charniak, 1996; Sima'an, 1996a). This is remarkable, since *DOP is not trained*: it reads the rules or elementary trees *directly* from hand-parsed sentences in a treebank, and calculates the probability of a new tree on the basis of *raw* subtree-frequencies in the corpus. Thus, one can seriously put into question the merits of sophisticated training or learning algorithms, such as Inside-Outside training (Pereira and Schabes, 1992) or Transformation-Based learning (Brill, 1993). And also Goodman's results (see his section "Experimental Results and Discussion"), support the hypothesis that an untrained grammar of treebank-subtrees outperforms a trained grammar of production rules.

---

[3] In my thesis, I did not calculate the *structural consistency*; I only calculated and reported the *accuracy* and *coverage* for the particular test set. Since DOP consists of an extremely overgenerating grammar, I also assumed that a 99% coverage means that for 99% of the sentences a perceived parse is among the found parses (cf. Bod, 1995a, p. 64), but this may not be true. I therefore do not criticize Goodman for the fact that he initially misinterpreted my 99% coverage, although I believe he could have corrected this in his final paper after that I had pointed out this mistake on the basis of his draft paper. Happily, he did correct this mistake at the conference presentation of his paper (Empirical Methods in Natural Language Processing Conference, May 17-18, 1996, Philadelphia, PA).